# Semiconductor Defect Detection by Hybrid Classical-Quantum Deep Learning

Yuan-Fu Yang
National Tsing Hua University
101, Section 2, Kuang-Fu Road,
Hsinchu, Taiwan R.O.C.
yfyangd@gmail.com

Min Sun
National Tsing Hua University
101, Section 2, Kuang-Fu Road,
Hsinchu, Taiwan R.O.C.
summin@ee.nthu.edu.tw

## Abstract

*With the rapid development of artificial intelligence and autonomous driving technology, the demand for semiconductors is projected to rise substantially. However, the massive expansion of semiconductor manufacturing and the development of new technology will bring many defect wafers. If these defect wafers have not been correctly inspected, the ineffective semiconductor processing on these defect wafers will cause additional impact to our environment, such as excessive carbon dioxide emission and energy consumption.*

*In this paper, we utilize the information processing advantages of quantum computing to promote the defect learning defect review (DLDR). We propose a classical-quantum hybrid algorithm for deep learning on near-term quantum processors. By tuning parameters implemented on it, quantum circuit driven by our framework learns a given DLDR task, include of wafer defect map classification, defect pattern classification, and hotspot detection. In addition, we explore parametrized quantum circuits with different expressibility and entangling capacities. These results can be used to build a future roadmap to develop circuit-based quantum deep learning for semiconductor defect detection.*

## 1. Introduction

The industry presents a paradox. The achievement of global climate goals will depend on semiconductors. They are an integral part of solar arrays, wind turbines, and electric vehicles. But chip manufacturing requires huge amounts of energy and water, and it emit a lot of exhaust gas during the production process. In addition to switching to renewables, chipmakers could also implement efficiencies in fabs to reduce their carbon footprint, including ineffective processing on the defect wafer.

In the semiconductor manufacturing, there are three main types of defects: wafer defect map, defect pattern, and hotspot. Wafer defect map is used to visualize the distribution of defect patterns and identify potential process and tool issues. Continuous-monitoring wafer defect maps are crucial for yield management because a sudden increase in the problematic patterns can be feedback to operation engineers to resolve the related issue. Generally, it is known that conventional defect patterns such as cluster, scratch, and ring are closely related to a certain type of process. For example, as shown in Figure 1, the defect map "Edge" is caused by the damaged low thermal mass (LTM) pad. LTM is an industrial grade self-regulating heating cable used for pipelines and chambers. Under the high temperature and high pressure of chemical vapor deposition (CVD) process, LTM pad will gradually get aged and cracked. Then, it will produce particles that fall on the edge of wafer surface. Another example "Cluster" is a mass of particles on the wafer surface during the Etching process. This defect is usually caused by the aged component of side O-ring. When this defect maps are detected, it means that the chamber needs to maintenance and replace aging parts with new ones.

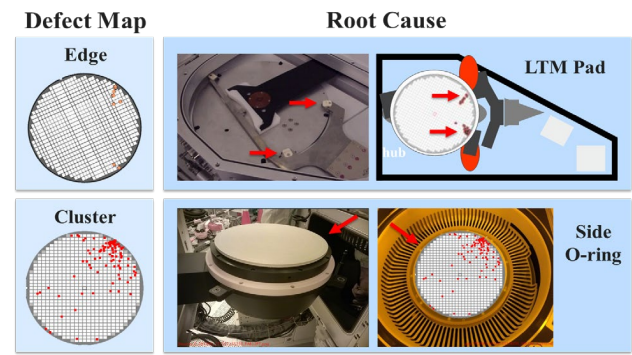

Figure 1: Defect map and its root cause.

The defect pattern is the unit that composes the wafer map, which image is obtained by a higher resolution inspection tool. Each defect pattern signals the root cause, and engineers can diagnose the failure and prevent it from happening again. As shown in Figure 2, the defect pattern "Multi-dots" mainly occurs in the WET Etching chamber pipeline and is caused by the dirty nozzles of xFlow. Another example is "Fallon" that often occurs on the CVD chamber. The main reason is the fall on particle caused by the damage of the point-of-use (POU) filter. The damage component must be replaced immediately to avoid this defect from happening again.

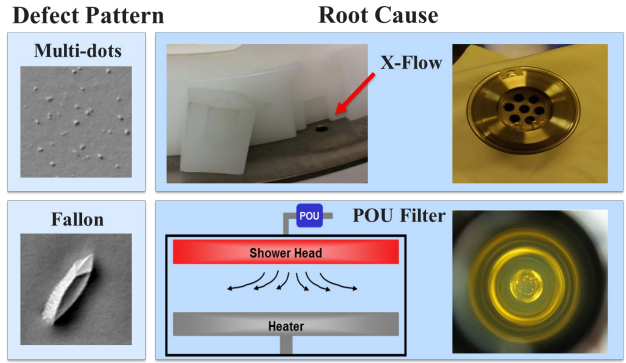

Figure 2: Defect pattern and its root cause

A hotspot is the region of mask layout patterns where failures including open and short circuits are more likely to happen during semiconductor manufacturing. Hotspot detection is used to check potential circuit failures at the post optical proximity correction (OPC) stage when transferring designed patterns onto silicon wafers. Figure-3 show the example of Epitaxy (EPI) hotspot and its root cause. We can find that this wafer has the layout pattern of SiGe wavy. This hotspot usually caused by tungsten loss (W loss) in the CVD process. In addition, the lithography (LIT) hotspot has received much attention in the recent years. LIT hotspots caused by light diffraction during manufacturing procedure due to a substantial mismatch between lithography wavelength and semiconductor technology feature size.

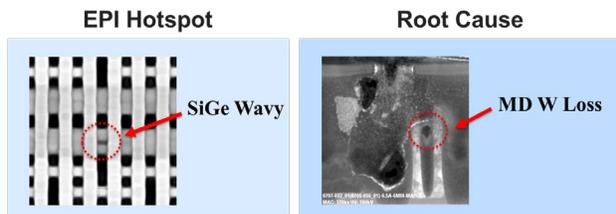

Figure 3: Hotspot and its root cause.

Deep learning as a hot computer vision technology has been actively realized in many fields. Although it has been serving the semiconductor industry in many areas, the contribution of deep learning in semiconductor defect detection is tremendous. However, deep learning algorithms tend to give probabilistic results and contain correlated components but at the same time suffer computational bottlenecks due to the curse of dimensionality. Similar to deep learning, quantum computing (QC) provides probabilistic results based on measurements formed by intrinsically coupled quantum systems. QC can provide potentially exponential speedups due to their ability to perform massively parallel computations on the superposition of quantum states.

In addition, each artificial neuron is usually constructed by linear connected layers, with non-linear activation functions connected at the end. In this work, we replace the linear part by the quantum circuit to take advantage of possible speedup in quantum computation. We propose the hybrid classical-quantum deep learning (HCQDL) based on quantum circuit for the above various types of defect inspection. Our HCQDL consists of classical and quantum layers. The classical layers implemented by the combination of self-proliferation and self-attention block. Self-proliferation block used a series of linear transformations to generate more feature maps at a cheaper computing cost. Then, self-attention block learned a wealth of information about long-range dependencies from this generated feature.

The quantum layer implemented by various quantum circuit built in the continuous-variable architecture, which encodes quantum information in continuous degrees of freedom such as the amplitudes of the electromagnetic field. By tuning parameters implemented on it, quantum circuit driven by our framework learns a given DLDR task, include of wafer defect map classification, defect pattern classification, and hotspot detection. The main contributions of the paper are:

(1) We introduce a new network architecture, Self-Proliferation-and-Attention block (SP&A Block), which can perform feature engineering in a more efficient way.

(2) We present the parametrized quantum circuit (PQC) with different expressibility and entangling capacities, and compare their training performance to quantify the expected benefits.

(3) We provide a future roadmap to develop circuit-based hybrid quantum-classical deep learning for semiconductor defect detection.

The rest of the paper has been organized as follows: Section 2 describes the related works of deep learning defect review (DLDR) and quantum machine learning. Section 3 presents our proposed HCQDL model and explore it in detail theoretically. Section 4 compares the results of latest DLDR models and our proposed method according to different metrics through several experiments. Finally, we conclude with some final thoughts and give future work recommendations in section 5.

## 2. Related works

In this section, we briefly discuss DLDR and quantum machine learning backgrounds.

### 2.1. Defect learning defect review

**Wafer Defect Map**. Wafer defect map is used to visualize defect patterns, identify potential process issues, and provide yield engineers with vital information to help them identify the root cause of die failures during semiconductor manufacturing processes. Nakata et al. [1] proposed the big data analysis enables comprehensive and long-term monitoring automation. They make use of fast and scalable methods of clustering and pattern mining and realize daily comprehensive monitoring with massive manufacturing data. They also apply deep learning to classification of wafer failure map patterns. Their deep learning model was a retrainable, one-class classifier with five layers whose parameters were determined empirically. They used to train it on the frequently occurring pattern, employing unlabeled images to monitor if that defect pattern emerges again. Nakazawa and Kulkarni [2] proposed the convolutional neural network (CNN) for wafer map defect classification and retrieval of similar maps. They demonstrate that by using only synthetic data for network training, real wafer maps can be classified with high accuracy. Kyeong and Kim [3] proposed a mixed defect classification system consisting of four CNN models for four basic defect types of the dataset. The proposed model was able to identify 16 defect types as combinations of four basic types. Kim et al. [4] presented a neural network-based bin coloring method and built a four-layered CNN to distinguish good and bad wafers. However, they did not classify the defect wafers into their respective defect types, which is a necessary step to analyze the root cause of the defects. di Bella et al. [5] adopted submanifold sparse CNN as a binary classifier for sparse data of larger images. They also used oversampling to overcome class imbalance, such as rotations, flips, and noise injection. Kong and Ni [6] proposed a multi-step detection system for multi-defect wafers classification. First, a binary CNN is used to classify wafers with overlapping and non-overlapping patterns. The wafer maps with a single pattern or non-overlapping mixed-type pattern were segmented into single pattern maps and then classified by a CNN. In the latest research, deep learning architecture has become the mainstream [8-10] on wafer defect map classification task. We plan to combine the advantages of deep learning and quantum computing as a roadmap for future research.

**Defect Pattern**. The defect pattern is the unit that composes the wafer map. It is the image obtained by a higher resolution inspection tool. Defect pattern recognition is suited to deep learning, a powerful supervised learning technique that does not need manual feature design. Beuth et al. [11] used a biological visual attention method to perform the deep learning defect review. This biological visual attention mechanism is mainly to realize the boundary search of the defect chip by simulating the signal of the pre-frontal cortex (PFC). However, the attention mechanism of this study has two problems: 1. The PFC of this model is only applicable to the search of geometric shapes, and not for textures. 2. The model cannot adaptively learn the attention area of attention. Therefore, how to develop an adaptive attention mechanism (self-attention) for defect pattern classification is the direction of future research. Chen et al. [12] proposed a lightweight network called WDD-Net. The method refers to MobileNet-V2, using depthwise separable convolutions to reduce parameters and calculations, The experimental results show that the detection speed of WDD-Net is 5 times faster than that of VGG-16. Yang and Sun [13] proposed a new DLDR architecture, named self-proliferating neural network (SPNet), which can effectively generate more feature maps with a lower computational cost.

**Hotspot**. Hotspot detection is defined as the procedure of finding the hotspots from the layout that would cause printability issues during lithograph. F. Yang et al. [14] used support vector machine (SVM) as the lithography hotspot classifier. They adopted spectral clustering for feature extraction, and their accuracy reached 95.66% on the public dataset ICCAD-2012. V. S. Ajna and N. George [15] proposed a method of layout hot spot detection based on deep learning. This method achieves 93% accuracy with the least number of false alarms. In subsequent research, deep learning became the main model architecture of hotspot detection [16-18]. X. Huang et al. [19] proposed an ensemble deep learning based on multiple sub-models. This method achieves recall rate 98.8% on the ICCAD-2012.

Quantum computation is a paradigm that furthermore includes nonclassical effects such as superposition, interference, and entanglement, giving it potential advantages over classical computing models. Therefore, we introduce the application of quantum machine learning in the next section, hoping to improve the weakness of classical deep learning by using quantum computing.

## 2.2. Quantum machine learning

Recent developments in quantum computing allowed scientists to look at computational problems from a new perspective. Researchers have been investigating quantum computing tools for a computational advantage for the deep learning problem. As near-term quantum devices move beyond the point of classical stimulability, also known as quantum supremacy [20], it is of utmost importance to discover new applications for noisy intermediate scale quantum (NISQ) devices [21] which are expected to be available in the next few years. Among the most promising applications for quantum computing is Quantum Machine Learning (QML) [22-23]. Recent advances in QML have been dominated by a class of algorithms called hybrid quantum-classical variational algorithms. Sasaki and Carlini [24] have delved into the notion of semi-classical and universal strategies whereby, in the former, classical methods are adapted to work on quantum systems whereas in the latter the methods are purely quantum in nature. A. Ajagekar and F. You [25] proposed the fault diagnosis deep learning method based on quantum computing. This method enjoys superior performance with an average fault diagnosis rate of 80% and tremendously low false alarm rates for the monitoring of Tennessee Eastman process.

In this paper, we present quantum computing based deep learning methods for semiconductor defect detection that are capable of overcoming the computational challenges faced by conventional techniques performed on classical computers. The proposed model effectively detects defect pattern by leveraging the superior feature extraction of self-proliferation and self-attention to facilitate proper classification between normal and defect wafers. Then, a quantum computing assisted generative training process followed by supervised discriminative training is used to train this model.

## 3. Method

Our technique, called hybrid classical-quantum deep learning (HCQDL), uses quantum circuits to nonlinearly transform classical inputs into features that can then be used in a number of deep learning algorithms. HCQDL consists of classical layer, quantum layer, and fully connection layer (as shown in Figure 4). The classical layer is implemented by the self-proliferation and self-attention (SP&A) block that used to extract feature maps effectively. The quantum layer is composed of the quantum circuit that can generate highly complex kernels whose calculations could be classically intractable. The fully connection layer is implemented by the non-linear activation functions to calculate the probability of various semiconductor defect. To demonstrate the power of the proposed method, we present results of using HCQDL to generate eigenstates of simple molecules, complex entangled ground states, and ground states of the transverse Hamiltonian with varying local fields and spin-spin couplings. The approach achieved high accurate results and can be generalized for creating quantum states of complex systems.

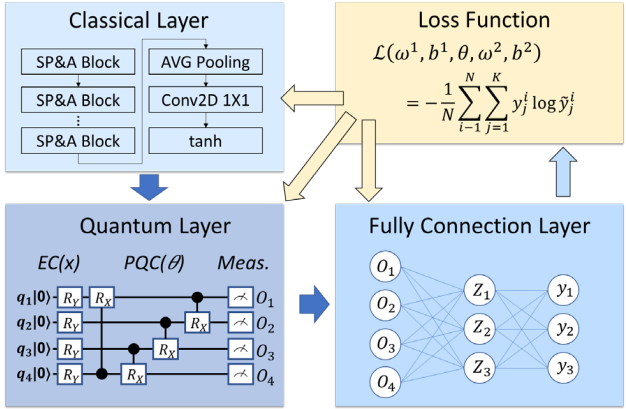

Figure 4: Architecture of hybrid classical-quantum deep learning.

### 3.1. Classical Layer

The classical layer is mainly composed of two parts: self-proliferation block and self-attention block. Self-proliferation block used a series of linear transformations to generate more feature maps at a cheaper cost. Self-attention block learned a wealth of information about long-range dependencies from this generated feature maps.

**Self-Proliferation Block.** Traditional convolution is a series of convolution operations to increase the feature depth. Self-proliferation block generates the same number of features through linear transform. The process is similar to DNA unzipping and replication, which can effectively increase the number of features. The process of Self-proliferation is as follows (as shown in Figure 5).

The first step of self-proliferation is a small amount of classical convolution which can be built upon a transformation $f \in \mathbb{R}^{c \times k \times k \times n}$ mapping an input $M \in \mathbb{R}^{c \times h \times w}$ to feature maps $M' \in \mathbb{R}^{h' \times w' \times n}$:

$$M' = M \otimes f \qquad (1)$$

where $\otimes$ denotes convolution, $c$ and $n$ is the number of input and out channels, $h$ and $h'$ is the height of the input and out data, $w$ and $w'$ is the width of the input and out data. we can obtain n feature maps by classical convolution.

The second step is a cheap operation, represented by $\rho$ in the Eq. 2. This is a linear transformation that uses depth-wise convolution to further obtain $n$ feature maps according to the following function:

$$u_{i,j} = \rho_{i,j}(m'_i), \forall i = 1, \dots, s, j = 1, \dots, t \qquad (2)$$

Where $m'_i$ is the $i$-th unit of feature map $M'$ generated by Eq. 1., $\rho_{i,j}$ is the $j$-th linear transformation for generating the $j$-th feature map $u_{i,j}$. If 12 feature maps are generated, self-proliferation can generate 6 of them (as shown in Figure 5). Therefore, the computation cost can be reduced by half.

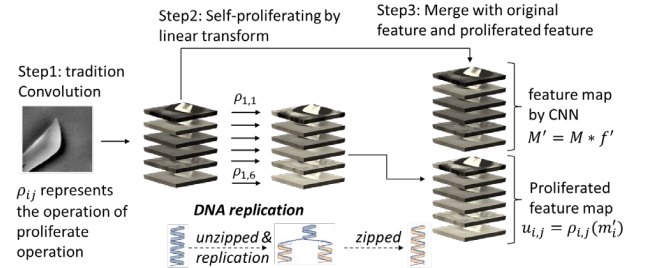

Figure 5: Self-proliferation process.

**Self-Attention Block.** The self-attention block aims at strengthening the features of the query position via aggregating information from other positions. The basic architecture formulated as:

$$y_i = x_i + w_{e2} ReLU \left( LN \left( w_{e1} \sum_{j=1}^{h \times w} \frac{e^{w_g x_j}}{\sum_{m=1}^{h \times w} e^{w_g x_m}} \right) \right) \quad (3)$$

We denote $x_i$ as the feature map of on input instance (e.g., defect pattern), where $h$ and $w$ is the height and width of input $x$. $w_{e1}$ and $w_{e2}$ denote linear transform matrices (1×1 convolution) which is used to bottleneck transform. $w_g$ is the weight for global attention pooling. *LN* denotes the layer normalization, which is utilized to filter the redundant information and refine the obtained contextual information. Specifically, our self-attention block consists of 3 parts (as shown in Figure 6): (a) global attention pooling for context modeling; (b) bottleneck transform to capture channel-wise dependencies; and (c) denotes the fusion function to broadcast element-wise addition for feature fusion. This function can aggregate the global context features to the features of each position [30].

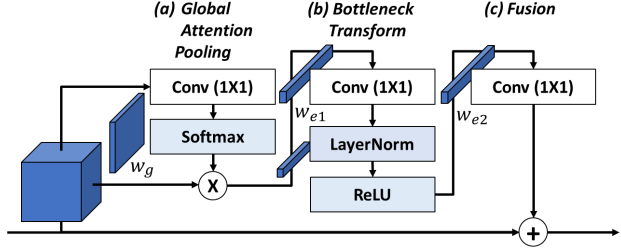

Figure 6: Architecture of self-attention.

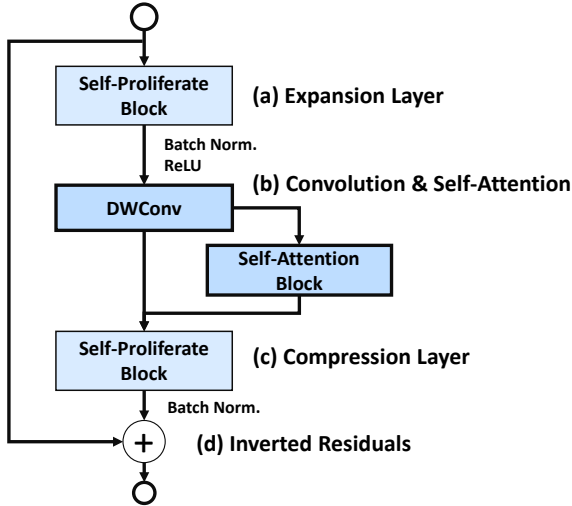

Figure 7: Architecture of self-proliferation-and-attention block.

**Self-Proliferation-and-Attention Block.** It consists of 4 parts (as shown in Figure 7): (a) expansion layer used self-proliferation block to increase input dimension to generate more feature maps at a cheaper cost. (b) depthwise convolution are used for feature extraction and self-attention are used to capture long-range dependencies. (c) compression layer used to reduce feature dimension to be the same as input feature. Then, we can transform spatial information with low computation. Finally, in (d) we refer to MobileNetV2 [26] to build inverted residuals. The residuals architecture was first proposed by He et al. in ResNet [27]. Its design concept is to learn residuals on the network to avoid gradient vanishing problems. ResNet's architecture is to compress first then expand. MobileNet's architecture is the opposite of ResNet that expand first then compress. The idea of MobileNet is to "capture features in high dimensions and transfer information in low dimensions".

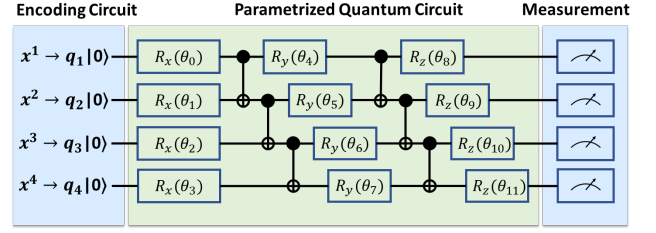

Figure 8: An example of quantum circuit with 4 qubits. Each qubit uses the rotation gate $R(\theta)$ by the angle $\theta$ around $x, y, z$. CNOT gate are used for every 2 qubits in order.

### 3.2. Quantum Layer

The quantum layer implemented by various quantum circuit built in the continuous-variable architecture. It consists of three consecutive parts (as shown in Figure 8). An encoding circuit encodes classical data to states of the qubits followed by a parametrized quantum circuit (PQC) that is applied to transform these states to their optimal location on the Hilbert space (as shown in Figure 9). Finally, measure the output of PQC along the z-axis with the $\sigma_z$ operator.

#### 3.2.1 Encoding Circuit

The encoding circuit is used to encode the classical data into the physical states of Hilbert space [28] for quantum computing. In this paper, we applied 3 encoding methods, including basis encoding, amplitude encoding, and angle encoding.

**Basis Encoding.** In basis encoding, the data has to be in form of a binary string to get encoding. Approximating a scalar value to its binary form and then transforming it to a quantum state. For example, if we have a classical dataset containing two examples $x_k = 01$ and $x_{k+1} = 11$, the corresponding quantum state after basis encoding is $|x_k\rangle = |01\rangle$ and $|x_{k+1}\rangle = |11\rangle$. In short, Basis encoding encodes an n-bit binary string $x_k$ to an $n$-qubit quantum state as:

$$|x_k\rangle = |i_x\rangle \quad (4)$$

where $|i_x\rangle$ is a computational basis state, and every binary string has a unique integer representation $i_x = \sum_{k=0}^{n-1} 2^k x_k$.

**Amplitude Encoding.** The amplitude encoding is also known as wave function embedding. The amplitude is the height of a wave. In the amplitude encoding, the data points are transformed into amplitudes of the quantum state. A normalized classical $N$-dimensional $x_k$ is represented by the amplitudes of a $n$-qubit quantum state $|x_k\rangle$ as:

$$|x_k\rangle = \sum_{k=0}^{N} x_k |i_x\rangle \quad (5)$$

where N is the length of vector $x_k$ into amplitudes of an n-qubit quantum state with $n = log_2(N)$. $\{|i_x\rangle\}$ is the computational basis for the Hilbert space. Since the classical information forms the amplitudes of a quantum state, the input needs to satisfy the normalization condition: $|x|^2 = 1$.

**Angle Encoding.** Angle encoding makes use of rotation gates to encode classical information $x_k \in \mathbb{R}^N$ without any normalization condition. The classical information determines angles of rotation gates:

$$|x_k\rangle = \otimes_{k=0}^{N} R(x_k)|0^N\rangle \qquad (6)$$

Where $N$ is the number of qubits, $R$ can be one of $R_x$, $R_y$, $R_z$. Usually, $N$ used for encoding is equal to the dimension of vector $x_k$. Since it is $2\pi$-periodic one may want to limit $\mathbb{R}^N$ to the hypercube $[0, 2\pi]^{\otimes n}$. The $i$-th feature $x_k$ is encoded into the $k$-th qubit via a Pauli-X rotation. Figure 9 show the angle encoding with a rotation angle of $\theta$ around $z$-axis on the Hilbert space. After activation function $tanh(x)$, the output $x_k \in [-1, 1]$ from the end of classical layer. Then, the rotation angle is mapped between $\theta \in [0, \pi]$ due to the periodicity of the cosine function. This is relevant since the expectation value is taken with respect to the $\sigma_z$ operator at the end of the circuit execution.

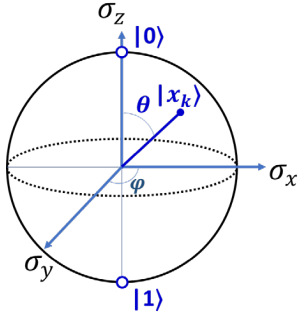

Figure 9: Angle encoding with a rotation angle of $\theta$ around $z$-axis on the Hilbert space.

### 3.2.2 Parametrized Quantum Circuit

A parameterized quantum circuit is a quantum circuit consisting of parameterized gates with fixed depth. These circuits have free parameters: the rotation angle of the quantum state. We use quantum circuits and repeat multiple times with different random parameters, instead of using the large and complex classical neural networks. The parameterized quantum circuit also consists of one-qubit gates as well as Controlled Not (*CNOT*). Some more complicated gates may also be used in PQC which can be decomposed into one qubit gates and *CNOT*. In general, an $n$ qubits PQC can be written as:

$$u(\hat{\theta})|\varphi\rangle = \left(\prod_{i=1}^{m} u_i\right)|\varphi\rangle \qquad (7)$$

where $u(\hat{\theta})$ is the set of unitary gates and m is the number of quantum gate. $\hat{\theta}$ is the set of parameters $\{\theta_0, \theta_1, ... \theta_k\}$, where $k$ is the total number of parameters and $|\varphi\rangle$ is the initial quantum state after data encoding. The unitary gate taking parameters is rotation gate $R(\hat{\theta})$, given by:

$$R_x(\hat{\theta}) = e^{-i\frac{\hat{\theta}}{2}\sigma_x}, R_y(\hat{\theta}) = e^{-i\frac{\hat{\theta}}{2}\sigma_y}, R_z(\hat{\theta}) = e^{-i\frac{\hat{\theta}}{2}\sigma_z} \qquad (8)$$

where $\{\sigma_x, \sigma_y, \sigma_z\}$ is Pauli matrices. The operation of $u$ can be modified by changing parameters $\hat{\theta}$. Thus, the output state can be optimized to approximate the wanted state by changing parameter $\hat{\theta}$. By optimizing the parameters used in $u(\hat{\theta})$, PQC approximates the wanted quantum states. To achieve better entanglement of the qubits before appending nonlinear operations, the $n$ qubits PQC has $n$ repeated layers in our model. By optimizing the parameters, the general PQC tries to approximate arbitrary states so that it can be used for different specific molecules.

In order to provide computational speedup by orchestrating constructive and destructive interference of the amplitudes in quantum computing, we constructed m rotation gates $R_x$ on the n qubits PQC as our basic quantum circuit, which can be written as:

$$\prod_{i=1}^{n} (\otimes_{j=0}^{m} R_x(\theta_{i+n\times j}) \, CNOT_{i,i+1}) \qquad (9)$$

where $R_x$ represents the unitary gate of rotation-$x$. $\theta_{i+n\times j}$ is adjustable parameter of unitary gates. $CNOT_{i,i+1}$ represents $CNOT$ gate with $m$ as the control qubit, and $n$ is the total number of qubits. Figure 10 show the basic quantum circuit with n=4 and m=3. Each qubit uses the rotation gate $R(\theta)$ by the angle $\theta$ around $x$-axis on the Hilbert space, and $CNOT$ gate is used for every 2 qubits.

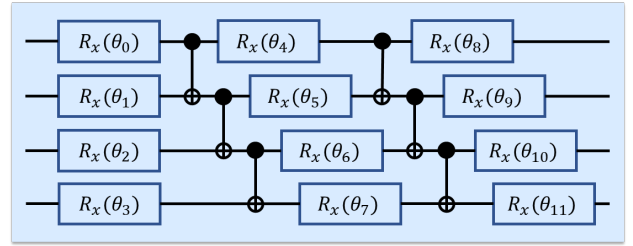

Figure 10: Basic quantum circuit with n=4, m=3

### 3.3. Fully Connection Layer

After obtaining all features from PQC, we feed them into a fully connection (FC) layer. We use the softmax activation function, so the output of FC layer will be a probability distribution. The i-th element of the output is the probability that this data point belongs to the i-th category, and we predict that this data point belongs to the category with the highest probability. In order to predict the actual label, we calculate the cumulative distance between the predicted label and the actual label as the loss function to be optimized:

$$\mathcal{L}(\omega^1, b^1, \theta, \omega^2, b^2) = -\frac{1}{N}\sum_{i=1}^{N}\sum_{j=1}^{K} y_j^i \log \tilde{y}_j^i \qquad (10)$$

where $\omega^1, b^1$ is the parameter of classical layer. $\theta$ is the parameter of quantum layer. $\omega^2, b^2$ is the parameter of FC layer. Perform the above procedure enough times to get a good estimate of the expected value of $\tilde{y}_j^i$ with minimum loss function.

## 4. Results

### 4.1. Experiment Circuits

In deep learning, it is very useful to transform data into a higher-dimensional feature space. In the same way, there are two strategies for using quantum circuits to generate higher dimensional features: entangling more and more qubits. However, quantum computers are now in their infancy, the available qubits are limited. In this experiment, we adopt different entangling strategies to verify our method on the 4 qubits circuit system. In addition to the basic quantum circuit (as shown in Figure-10), we also chose Circuit-5/6/16/17 provided by Sim et al. [29], which has better expressive ability and entanglement ability, as the parametrized quantum circuit (as shown in Figure 11).

Table 1: Ablation Study of Defect Pattern Classification

| Model | Classical Layer Feature Extraction | Quantum Layer | | | | | |
|---|---|---|---|---|---|---|---|
| | | Encoding | Circuit | | | | |
| | | | Basic | 5 | 6 | 16 | 17 |
| Hybrid | ResNet50 [27] | Basic | 93.73 | 95.42 | 95.03 | 94.42 | 93.94 |
| | | Amplitude | 93.74 | 95.43 | 95.04 | 94.43 | 93.95 |
| | | Angle | 93.81 | 95.50 | 95.11 | 94.51 | 94.02 |
| Classical | | - | 92.76 | | | | |
| Hybrid | SENet [45] | Basic | 95.09 | 96.75 | 96.43 | 96.04 | 95.85 |
| | | Amplitude | 95.27 | 96.94 | 96.61 | 96.22 | 96.03 |
| | | Angle | 95.42 | 97.08 | 96.75 | 96.36 | 96.17 |
| Classical | | - | 93.55 | | | | |
| Hybrid | MobileNetV3 [31] | Basic | 92.93 | 94.13 | 93.70 | 93.24 | 93.40 |
| | | Amplitude | 93.02 | 94.21 | 93.79 | 93.33 | 93.49 |
| | | Angle | 93.39 | 94.58 | 94.15 | 93.70 | 93.86 |
| Classical | | - | 92.09 | | | | |
| Hybrid | SP&A-Net (proposed model) | Basic | 95.92 | 97.85 | 97.05 | 96.54 | 96.11 |
| | | Amplitude | 96.06 | 97.99 | 97.19 | 96.68 | 96.25 |
| | | Angle | 96.55 | 98.47 | 97.68 | 97.17 | 96.73 |
| Classical | | - | 94.27 | | | | |

### 4.2. Ablation Study

To demonstrate the usefulness of quantum circuit, our model had been verified on two industry data sets, defect pattern and EPI hotspot. We performed a rigorous ablation study and showed the quantitative comparison in Table 1. We refer to ResNet50, SENet, MobileNetV3 and our proposed model as the basic model to integrate with various quantum circuits. In addition to the basic circuit we proposed, we added the Circuit-5/6/16/17 presented in the previous section. We also adopted four encoding strategy: basic, amplitude, and angle encoding.

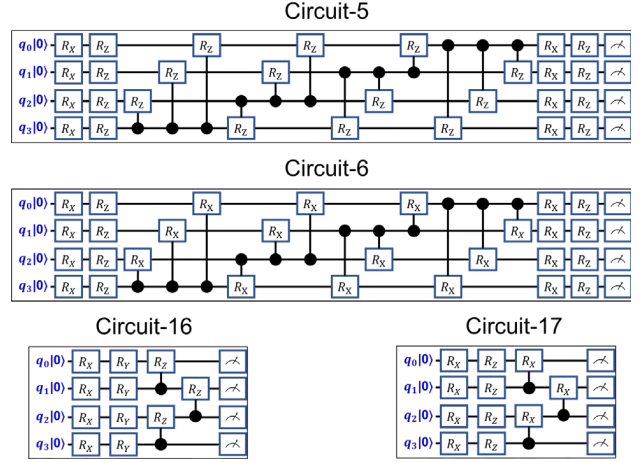

Figure 11: Quantum Circuit by Sim et al. [29].

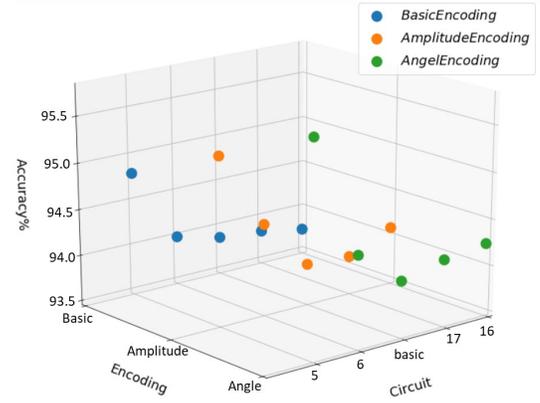

Figure 12: ResNet50 [27] with Different Strategy of Quantum Circuit for Defect Pattern Classification

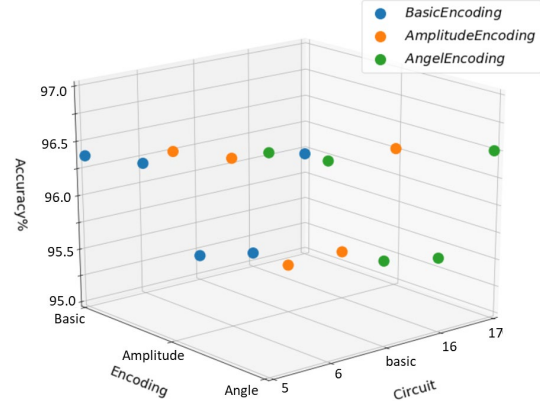

Figure 13: SENet [45] with Different Strategy of Quantum Circuit for EPI Hotspot Classification

Ablation results yielded many significant findings. First, Circuit 5 and Circuit 6 is a fully connected graph arrangement of qubits which led to both favorable expressibility and entangling capability. Therefore, the network model with Circuit 5 and Circuit 6 has high accuracy regardless of semiconductor defect detect task (as shown in Figure 12 and Figure 13). Second, Circuit 5 and Circuit 16 with controlled Z-rotation ($CRz$) gates is better than Circuit 6 and 17, respectively. It because that $CRz$ operations in the entangling block commute with each other and thus the effective unitary operation comprised of $CRz$ gates can be expressed using unique generator terms that are fewer than the number of parameters for these gates. Finally, the basic circuit without any controlled rotation gate has the lowest accuracy among the 4 classification tasks. This suggests that, if one is trying to design a PQC to increase expressibility, it is better to insert single-qubit gates which skew the controlled gate rotation axis away from the control axis.

### 4.3. Experiment Results

We compare our proposed model on the public dataset of LIT hotspot (ICCAD-2012) and defect wafer map (WM-811K) with the state-of-art method in the last 4 years. Table 2 and table 3 shows that our proposed model (HCQDL) has a higher test accuracy, which shows the advantages of quantum computing. It mean that small quantum circuits provide significant performance lift over standard linear classical algorithms, improving classification accuracy rates from around 98% to 99.12%, and from around 96% to 98.10% in these two public dataset, respectively.

Table 2: Comparison of state-of-the-art methods on ICCAD-2012

| Authors | Model | Feature Extraction | Data Augmentation | Accuracy |
|---|---|---|---|---|
| H. Yang, et al. [32] | CNN | Feature Tensor Extraction | Mirroring, Flipping | 98.88% |
| H. Yang, et al. [33] | CNN | - | Mirroring, Flipping, Upsampling | 98.20% |
| Y. Tomioka, et al. [34] | Real AdaBoost | Histogram of Oriented Light Propagation | Rotating, Reflecting | 99.01% |
| F. Yang, et al. [35] | Efficient SVM | Spectral Clustering | - | 95.66% |
| V. Borisov and J. Scheible [36] | CNN | - | Flipping, Gaussian filter | 98.94% |
| H. Yang, et al. [37] | CNN | - | - | 97.36% |
| V.S. Ajna and N. George [15] | CNN | - | - | 93.00% |
| X. Huang, et al. [19] | Ensemble CNN | Combine Physical Features | - | 98.80% |
| X. Lin, et al. [38] | Heterogeneous Federated Learning | Feature Tensor Extraction | - | 97.90% |
| Our Proposed Model | Hybrid Classical-Quantum CNN | Self-Proliferate Self-Attention | Mirroring, Flipping, Rotating | **99.12%** |

Table 3: Comparison of state-of-the-art methods on WM-811K

| Authors | Model | Feature Extraction | Data Augmentation | Accuracy |
|---|---|---|---|---|
| T. Nakazawa and D. V. Kulkarni, [2] | CNN | - | - | 96.30% |
| N. Yu, Q. Xu and H. Wang [7] | CNN | PCA | - | 93.25% |
| J. Yu and J. Liu [10] | CNN | AE+PCA | - | 97.30% |
| J. Yu, X. Zheng and J. Liu [39] | CNN | AE | - | 95.13% |
| J. Yu [40] | SVM | AE | - | 89.50% |
| T.-H. Tsai and Y.-C. Lee [8] | MobileNetV2 | - | AE | 97.01% |
| M. B. Alawieh et al. [9] | CNN | - | AE | 94.00% |
| S. Kang [41] | CNN | - | Rotation | 92.13% |
| U. Batool et al. [42] | CNN | - | Flipping, Brightness | 90.44% |
| M. Saqlain et al. [43] | CNN | - | Rotation, flipping, shifting, zooming | 96.20% |
| D. Kim and P. Kang [44] | CNN | - | - | 96.40% |
| Our Proposed Model | Hybrid Classical-Quantum CNN | Self-Proliferate Self-Attention | Mirroring, Flipping, Rotating | **98.10%** |

## 5. Conclusion

Our experiments in this work were designed to highlight the novelties introduced by the quantum deep learning: the generalizability of convolutional layers inside a typical CNN architecture, the ability to use this quantum circuit on semiconductor defect datasets, and the potential use of features introduced by the parametrized quantum circuit. The experimental results show that our framework outperforms the existing deep learning techniques. In addition, based on recent progress, we reasonably believes that quantum advantage is achievable within the next 5-10 years. Quantum computers promise access to fast linear algebra processing capabilities which are in principle able to deliver the polynomial speed-up that allows kernel methods to process big data without relying on approximations and heuristics.

Moreover, there is much to be learned about how to select these parameterized circuits, how to choose the random numbers to mix with the input data, and how to optimize the number of qubits to get good performance. Many of our existing deep learning algorithms could be translated into hybrid classical-quantum deep learning, allowing them take advantage of new properties like entanglement, superposition and parallel computing. It's only a matter of time before these algorithms are changing the world. We hope this paper can be a demonstration that quantum machine learning has the potential to provide satisfactory solutions for semiconductor defect detection.